\documentclass{article}

\usepackage[T1]{fontenc}
\usepackage[utf8]{inputenc}
\usepackage{latexsym}
\usepackage{microtype}
\usepackage{graphicx}
\usepackage{hyperref}

\usepackage[preprint]{icml2026}
\hypersetup{pdfsubject={arXiv preprint}}
\usepackage{xurl}
\usepackage{amsmath,amssymb,booktabs,tabularx,multirow}
\usepackage{placeins}

\newcommand{\Require}{\REQUIRE}
\newcommand{\State}{\STATE}
\newcommand{\For}{\FOR}
\newcommand{\EndFor}{\ENDFOR}
\newcommand{\Return}{\textbf{return}\ }

\definecolor{improvementgreen}{RGB}{0,128,64}
\definecolor{teachergray}{gray}{0.45}
\newcommand{\SDOPD}{\textsc{SD-OPD}}
\newcommand{\Frozen}{\textsc{Frozen}}
\newcommand{\Ungated}{\textsc{Ungated}}
\newcommand{\sg}{\operatorname{sg}}
\newcommand{\ind}{\mathbf{1}}
\newcommand{\cent}{\mathcal{C}}
\newcommand{\approxcent}{\widetilde{\mathcal{C}}}

\newcommand{\StyleAmplitude}{1.226}
\newcommand{\MathAmplitude}{0.347}
\newcommand{\AmplitudeRatio}{3.5}

\newcommand{\GateStyleReduction}{69.1}
\newcommand{\StudentBaseStyleMathRatio}{4.1}
\newcommand{\GatedStyleMathRatio}{3.1}
\newcommand{\GatedOPDStyleAmplitude}{0.380}
\newcommand{\GatedOPDMathAmplitude}{0.102}
\newcommand{\GatedStudentMathAmplitude}{0.074}
\newcommand{\GatedOPDStyleMathRatio}{3.7}
\newcommand{\GatedAmplitudeStyleReduction}{40.3}
\newcommand{\ObservationResponseCount}{8,192}

\newcommand{\StyleTokenMillions}{0.6}
\newcommand{\MathTokenMillions}{4.7}

\newcommand{\OPDAverage}{42.87}
\newcommand{\MainGain}{9.77}

\newcommand{\LargeTeacherSmallStudentGain}{9.99}

\newcommand{\MiMoOPDAverage}{50.65}
\newcommand{\MiMoExOPDAverage}{44.40}
\newcommand{\MiMoSDOPDAverage}{59.29}
\newcommand{\MiMoGain}{8.64}
\newcommand{\MiMoExOPDGap}{14.89}
\newcommand{\MiMoExOPDDrop}{6.25}

\newcommand{\FourBGain}{7.76}
\newcommand{\FourBExOPDGap}{3.45}

\newcommand{\EightBSDOPDAverage}{76.11}
\newcommand{\EightBGain}{10.37}
\newcommand{\EightBExOPDGap}{9.07}
\newcommand{\LargeTeacherAverage}{73.82}
\newcommand{\EightBTeacherGain}{2.29}

\newcommand{\FrozenAverage}{34.47}
\newcommand{\FrozenGap}{18.18}

\newcommand{\EvaluationResponseCount}{10,560}
\newcommand{\UngatedMeanLength}{6.05}
\newcommand{\UngatedMinLength}{5}
\newcommand{\UngatedMaxLength}{13}
\newcommand{\UngatedSixTokenCount}{10,249}
\newcommand{\MainSettingCount}{5}
\newcommand{\MainComparisonCount}{30}

\newcommand{\ExOPDBenchmarkWins}{28}
\newcommand{\MinOPDSettingGain}{7.76}
\newcommand{\MaxOPDSettingGain}{10.37}
\newcommand{\MinExOPDSettingGain}{3.45}
\newcommand{\MaxExOPDSettingGain}{14.89}
\newcommand{\PreferenceTryRetained}{19.3}
\newcommand{\PreferenceStrategyRetained}{13.8}
\newcommand{\PreferenceIdeaRetained}{71.4}

\icmltitlerunning{Reinforcing What the Student Has Learned}

\begin{document}
\raggedbottom
\twocolumn[
  \icmltitle{Reinforcing What the Student Has Learned: Teacher-Guided \texorpdfstring{\\}{ } On-Policy Distillation}
  \icmlsetsymbol{equal}{*}
  \icmlsetsymbol{corresponding}{\textdagger}
  \begin{icmlauthorlist}
    \icmlauthor{Bozhen Fang}{equal,zju}
    \icmlauthor{Linhao Zhong}{equal,zju,ant}
    \icmlauthor{Xuehao Wang}{zju}
    \icmlauthor{Huanyi Zheng}{zju}
    \icmlauthor{Hao Chen}{zju}
    \icmlauthor{Chunhua Shen}{zju,corresponding}
  \end{icmlauthorlist}
  \icmlaffiliation{zju}{Zhejiang University, State Lab of CAD \& CG}
  \icmlaffiliation{ant}{Ant Group}
  \icmlcorrespondingauthor{Chunhua Shen}{chunhuashen@zju.edu.cn}
  \vskip 0.2in
]
\printAffiliationsAndNotice{\icmlEqualContribution\unskip. \textsuperscript{\textdagger}Corresponding author. }

\begin{abstract}
On-policy distillation (OPD) improves language-model reasoning through dense teacher feedback on student-generated responses.
We first observe an amplitude disparity in this feedback: on math reasoning responses, style-related positions exhibit \AmplitudeRatio{} times the mean absolute teacher--student log-probability difference of mathematical positions.
Scaling OPD updates by these discrepancies thus directs much of the update strength toward matching the teacher's expression.
We instead take the student's own pretrained base as a reference: the student--base difference records the token preferences acquired during post-training and provides supervision that does not depend on the student's distance from the teacher.
Building on this reference, we propose Student-Delta On-Policy Distillation (\SDOPD{}), which uses centered student--base differences as update coefficients, keeps each coefficient only when its sign agrees with centered teacher--student feedback, and recomputes the student signal every training round.
Across \MainSettingCount{} teacher--student configurations with Qwen3 students of 1.7B to 8B parameters, \SDOPD{} improves six-task mathematics average pass@1 (avg@16) over OPD by \MinOPDSettingGain{} to \MaxOPDSettingGain{} percentage points.

Code is available at \url{https://github.com/aim-uofa/sd-opd}.

\end{abstract}
\begin{figure*}[!t]
 \centering
 \includegraphics[width=\textwidth]{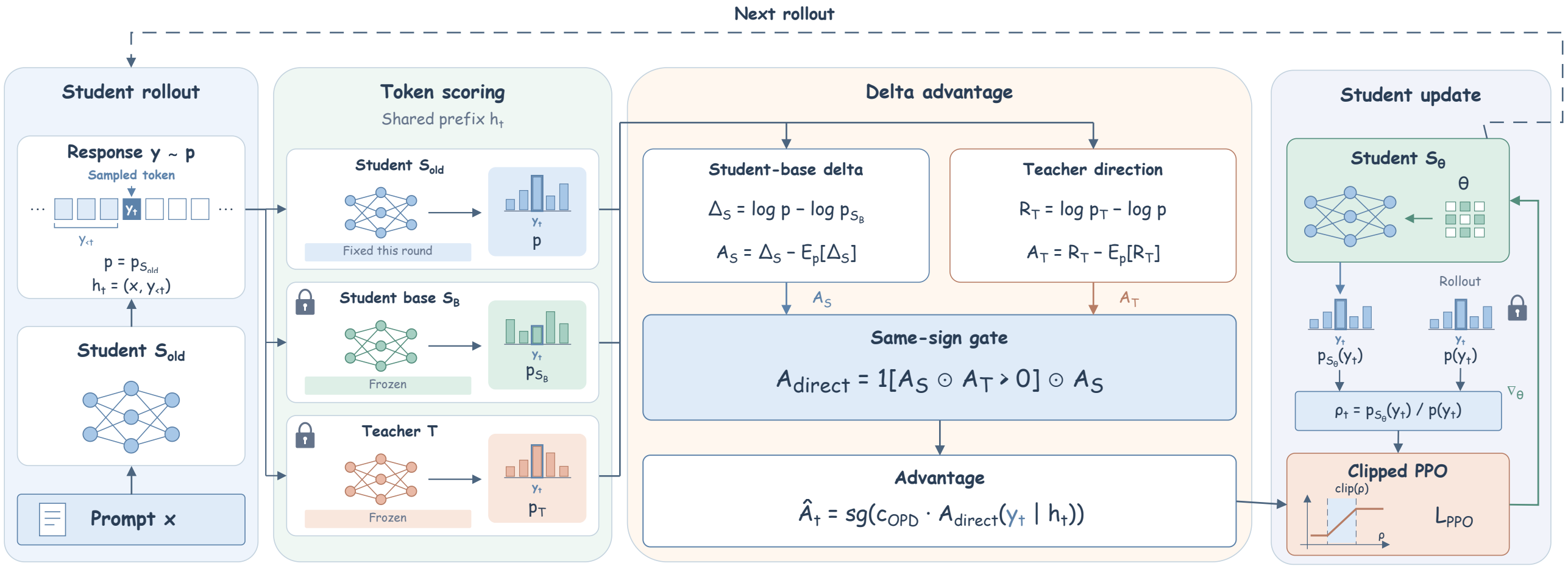}
 \caption{\SDOPD{} overview. Each round scores student-generated responses on shared prefixes. The student--base difference supplies the coefficient, and teacher--student feedback selects its admissible direction. Here, $A_T=A_{\mathrm{OPD}}$, $A_{\mathrm{direct}}=A_{\text{SD-OPD}}$, and $\rho_t=w_t$. The diagram shows full-vocabulary centering; training uses the top-1,024 approximation in Appendix~\ref{app:candidate-centering}. Advantages remain fixed within each round and are recomputed after refreshing the student. Teacher $T$ and base $S_B$ remain frozen.}
 \label{fig:method}
\end{figure*}

\section{Introduction}
On-policy distillation (OPD) is a post-training approach in which a language model learns from a teacher on responses generated by its own current policy \citep{agarwal2024policy,gu2023minillm}.
The teacher scores every token at prefixes the student visits, so supervision is dense and lies on the student's own trajectories.
As the student changes, new responses are sampled and scored, and the training distribution follows the student's current behavior, as in imitation learning on learner-induced states \citep{ross2011reduction}.
This combination of on-policy sampling and token-level feedback has supported both capability transfer from larger teachers to smaller students \citep{yang2025qwen3,yang2026learning} and the consolidation of specialized teachers into a single model \citep{xiao2026mimo}.

However, the effectiveness of OPD depends on how teacher feedback is constructed and weighted \citep{fu2026revisiting,wang2026dr}.
In sampled-token OPD, the magnitude of teacher--student log-probability differences determines the strength of token-level supervision \citep{gu2023minillm,yang2026learning}.
Post-training also shifts a model's preferences for wording and formatting \citep{lin2024unlocking,zhou2023lima}, so teacher--student disagreement reflects differences in expression as well as gaps in reasoning.
Weighting updates by this disagreement thus spends update strength on the teacher's expression as well as its reasoning.
This raises two questions: how is update strength distributed across tokens, and can it come from a source other than teacher disagreement?

\begin{figure*}[t]
 \centering
 \includegraphics[width=0.98\textwidth]{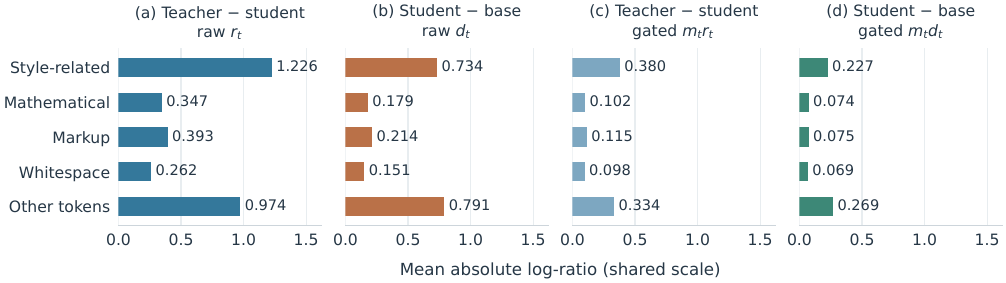}
 \caption{Feedback on the same student responses: raw (a) teacher--student $r_t$ and (b) student--base $d_t$, and both after the same-sign mask $m_t=\ind[r_td_t>0]$, (c) $m_tr_t$ and (d) $m_td_t$. Panels (c) and (d) retain identical positions and directions and differ only in magnitude source. All panels share the scale, categories, and averaging procedure in Equation~\ref{eq:amplitude}; filtered-out positions contribute zero and remain in the denominator.}
 \label{fig:amplitude}
\end{figure*}

\begin{figure*}[t]
 \centering
 \includegraphics[width=0.98\textwidth]{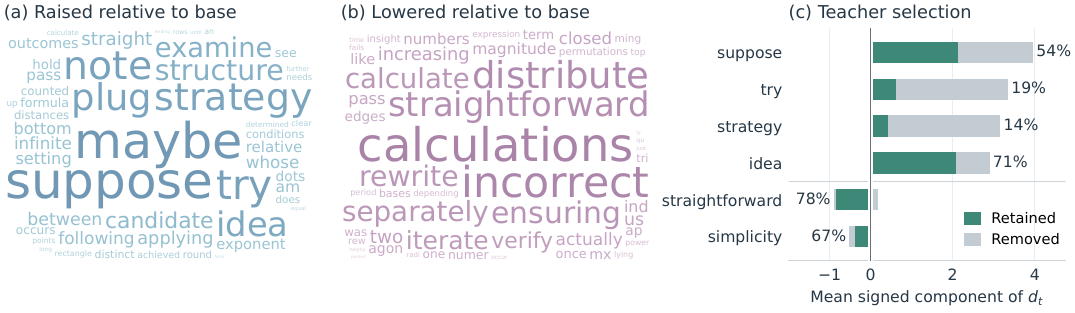}
 \caption{Student--base preferences and teacher selection. (a, b) Fifty representative strings per sign of $d_t$, with larger, darker text indicating greater averaged amplitude. A piecewise-linear size scale separates ten high-amplitude examples from smaller context strings. (c) Six representative strings: parts of $d_t$ with $d_t>0$ (right of zero) and $d_t<0$ (left), split into the part retained by $\ind[r_td_t>0]$ and the part removed, with the retained share of absolute amplitude. Appendix~\ref{app:preference-statistics} describes aggregation and string selection.}
 \label{fig:preferences}
\end{figure*}

For the first question, on DAPO-Math responses, style-related positions have \AmplitudeRatio{} times the mean absolute teacher--student log-probability difference of mathematical positions (Figure~\ref{fig:amplitude}).
For the second, we compare the student with its own pretrained base \citep{lin2024unlocking}, which records the token preferences the student has already acquired, independently of its distance from the teacher.
These preferences include reasoning expressions such as \emph{suppose}, \emph{try}, and \emph{strategy}, but their amplitude is also concentrated at style-related positions, with a style-to-mathematics ratio of \StudentBaseStyleMathRatio{}, so they need selection.
We select them by teacher sign agreement, keeping a student preference only where the teacher favors the same direction.
This test retains very different shares of preferences with similar strength (Figure~\ref{fig:preferences}) and lowers the style-to-mathematics ratio to \GatedStyleMathRatio{}, whereas the same test with teacher--student magnitudes still leaves \GatedOPDStyleMathRatio{} (Section~\ref{sec:content}).

These findings lead to \emph{Student-Delta On-Policy Distillation} (\SDOPD{}; Figure~\ref{fig:method}).
Centered student--base log-probability differences supply the coefficients, and a coefficient is kept only when its sign agrees with centered teacher--student feedback.
The teacher thus decides which student preferences to develop, but its disagreement magnitude does not scale the updates.
The teacher and base stay frozen, while the student signal is recomputed after each training round, so it covers both the initial post-training changes and those acquired during distillation.

We evaluate \SDOPD{} in \MainSettingCount{} teacher--student configurations with three teachers and Qwen3 students of 1.7B, 4B, and 8B parameters, using six mathematics benchmarks and matched rollout and response budgets.
\SDOPD{} outperforms OPD on every benchmark in every configuration and achieves the highest six-task student average in all \MainSettingCount{} configurations (Table~\ref{tab:main}).
Average gains range from \MinOPDSettingGain{} to \MaxOPDSettingGain{} percentage points over OPD and from \MinExOPDSettingGain{} to \MaxExOPDSettingGain{} over ExOPD.

Our contributions are as follows:
{\setlength{\topsep}{3pt}\setlength{\partopsep}{0pt}
\begin{itemize}
 \setlength{\itemsep}{3pt}\setlength{\parsep}{0pt}\setlength{\parskip}{0pt}
 \item \textbf{A diagnosis of how OPD weights its updates.}
 We show that teacher--student disagreement is not a neutral measure of what the student should learn: its mean absolute magnitude is \AmplitudeRatio{} times larger at style-related than at mathematical positions, the gap persists with a larger teacher and on GSM8K, and sign-agreement filtering does not remove it while teacher--student differences set the magnitude.
 \item \textbf{The student's own post-training changes as a source of supervision.}
 We show that the difference between a post-trained student and its pretrained base can set update strength, with the teacher deciding only which directions to reinforce.
 \SDOPD{} implements this division of roles and recomputes the student signal every round, so update strength follows what the current student prefers rather than how far it is from the teacher.
 It improves over OPD in every evaluated configuration and lets an 8B student exceed the average accuracy of its Qwen3-30B-A3B teacher.
\end{itemize}}

\section{Related Work}
\paragraph{On-policy knowledge distillation.}
Knowledge distillation transfers teacher predictions to a student \citep{hinton2015distilling}, while sequence-level distillation trains on teacher-generated targets \citep{kim2016sequence}.
Teacher-generated reasoning data extend this approach to mathematical and code reasoning \citep{guo2025deepseek,guha2026openthoughts}.
GKD and MiniLLM obtain supervision on student-generated sequences through distribution matching and reverse-KL policy optimization \citep{agarwal2024policy,gu2023minillm}.
DistiLLM combines skew-divergence objectives with reuse of student outputs, while DistiLLM-2 contrasts teacher- and student-generated responses \citep{ko2024distillm,ko2025distillm}.
Black-box OPD learns feedback from a discriminator trained on teacher and student text \citep{ye2025black}.
Recent work also studies local distribution matching and token-specific supervision weights \citep{fu2026revisiting,wang2026dr}.
In these methods, teacher outputs determine the strength of supervision; \SDOPD{} instead takes update strength from the student's own post-training changes and uses the teacher only to select their direction.

\paragraph{Self-distillation and contextual supervision.}
Context-conditioned self-teaching obtains additional information from reference solutions in Self-Distilled Reasoner \citep{zhao2026self}, textual feedback in SDPO \citep{hubotter2026reinforcement}, and demonstrations in SDFT \citep{shenfeld2026self}.
On-Policy Context Distillation also learns from a context-conditioned teacher, including configurations with different teacher and student sizes \citep{ye2026policy}.
In these approaches, the additional supervision comes from privileged information placed in the teacher's context; in \SDOPD{}, it comes from the student's own post-training history, measured against a fixed base model, with an independent teacher constraining its direction.

\paragraph{Post-training differences and delta distillation.}
URIAL analyzes changes between aligned and base token distributions, including changes in expression \citep{lin2024unlocking}; we turn this comparison into a training signal.
Policy--reference log-ratios connect preference optimization to implicit rewards \citep{rafailov2023direct}; implicit process reward models and PRIME extend this connection to dense reasoning feedback learned from outcome labels \citep{yuan2024free,cui2025process}.
G-OPD introduces a reference policy and reward scaling; its ExOPD variant studies extrapolation with both an initial-student reference and a teacher pre-RL reference \citep{yang2026learning}.
OPD$^2$ also uses centering and a same-sign condition, but its primary signal is a teacher--base difference \citep{heo2026policy}.
\SDOPD{} instead uses the current student's changes relative to the student's own base, recomputed every round, and lets teacher feedback decide which directions to reinforce.

\section{Observations}
\label{sec:observations}
We compare teacher--student disagreement with the student's own post-training changes, using the initial student $S_0$ (Qwen3-1.7B), its pretrained base $S_B$ (Qwen3-1.7B-Base), and teacher $T$ (Qwen3-4B-Instruct-2507).
All analyses use all eight responses to each of the same 1,024 questions sampled without replacement from DAPO-Math-17K \citep{yu2026dapo}.
This gives \ObservationResponseCount{} responses with about 11.3 million response tokens.
All three models then score the same token IDs at the same causal prefixes, using full-vocabulary log-softmax normalization.
Appendix~\ref{app:token-categories} gives generation and scoring details, and Appendix~\ref{app:observation-generalization} repeats the category-level analyses with another teacher and on GSM8K.

\subsection{Teacher--Student Disagreement Varies across Token Categories}
\label{sec:amplitude}
To compare teacher--student disagreement across token categories, we use the sampled-token log-ratio feedback studied in OPD \citep{gu2023minillm,yang2026learning}.
For response token $y_t$ at prefix $h_t=(x,y_{<t})$, this feedback is
\begin{equation}
 r_t=\log p_T(y_t\mid h_t)-\log p_{S_0}(y_t\mid h_t).
 \label{eq:rawopd}
\end{equation}
We categorize token positions using fixed textual rules.
Style-related positions include discourse wording, answer labels, and Markdown formatting, while mathematical positions include numbers, operators, formula variables, a fixed mathematical lexicon, and logical expressions.
Markup, whitespace, and other tokens form separate groups; Appendix~\ref{app:token-categories} details the classification.

Because the categories differ in frequency (about \StyleTokenMillions{} million style-related and \MathTokenMillions{} million mathematical tokens, both present in every response), we compare per-token means rather than totals.
For positions $\mathcal{T}_{i,j,c}$ in category $c$ of response $j$ to question $i$, we compute the per-token mean
\begin{equation}
 M_{i,j,c}(r)=\frac{1}{|\mathcal{T}_{i,j,c}|}
       \sum_{t\in\mathcal{T}_{i,j,c}}|r_{i,j,t}|.
 \label{eq:amplitude}
\end{equation}
We average these values equally over the eight responses to each question and then equally over questions, preventing longer responses from receiving extra weight.

\paragraph{Style-related positions have larger feedback amplitudes.}
Figure~\ref{fig:amplitude}(a) shows mean absolute feedback of \StyleAmplitude{} at style-related positions and \MathAmplitude{} at mathematical positions, a \AmplitudeRatio{}-fold difference.
OPD uses this log-ratio as a coefficient in token-level policy gradients \mbox{\citep{gu2023minillm,yang2026learning}}, so for the initial student, style-related positions receive larger coefficients on average.
For example, differences in wording or formatting receive large coefficients even when the student expresses the same mathematical idea.

\subsection{Student Preferences and Teacher Selection}
\label{sec:content}
Since teacher disagreement mixes expression with reasoning, we look for a reference that measures update strength from what the student has already learned.
The student's own pretrained base provides one: comparing student and base predictions at shared prefixes reveals preferences acquired during post-training, including how the student expresses mathematical reasoning \citep{lin2024unlocking}.
On the same responses and prefixes, we define
\begin{equation}
 d_t=\log p_{S_0}(y_t\mid h_t)-\log p_{S_B}(y_t\mid h_t).
 \label{eq:rawstudent}
\end{equation}
The sign of $d_t$ indicates whether the observed token has become more probable ($d_t>0$) or less probable ($d_t<0$) relative to the base at that prefix.
Since the reference is the pretrained base, this difference summarizes cumulative changes across post-training.

\paragraph{Acquired preferences favor exploration and planning.}
Figure~\ref{fig:preferences}(a--b) shows the strings whose probability post-training raised or lowered most.
The raised strings are reasoning moves rather than formatting: they form hypotheses and explore alternatives (\emph{suppose}, \emph{maybe}, \emph{try}), plan and organize a solution (\emph{strategy}, \emph{idea}, \emph{structure}), and test candidates (\emph{plug}, \emph{examine}).
The lowered strings instead narrate routine computation (\emph{calculations}, \emph{distribute}, \emph{rewrite}) or assert that a step is easy or correct (\emph{straightforward}, \emph{ensuring}), and their shifts are much smaller.
Post-training has thus mainly added ways of deciding how to approach a problem, rather than removing old habits of describing computations.
These changes reflect how this particular student learned to reason, and a teacher trained differently need not share them.
The student--base difference therefore records reasoning behavior the student has already acquired, which makes it a natural source of supervision for further distillation.
Not every shift is desirable, however: the checking step \emph{verify} is also lowered, so these changes still need selection.

\paragraph{Teacher agreement selects which acquired preferences to reinforce.}
These changes also include stylistic preferences, with a style-to-mathematics ratio of \StudentBaseStyleMathRatio{} (Figure~\ref{fig:amplitude}(b)).
We therefore keep $d_t$ only where the teacher agrees in sign, using the mask $m_t=\ind[r_td_t>0]$.

This test separates preferences of similar strength (Figure~\ref{fig:preferences}(c)): \emph{try} and \emph{strategy} keep only \PreferenceTryRetained{}\% and \PreferenceStrategyRetained{}\% of their absolute amplitude, whereas \emph{idea} keeps \PreferenceIdeaRetained{}\%.
For \emph{straightforward}, it keeps the decrease and removes occurrences where the student raised its probability.
Across categories, it removes \GateStyleReduction{}\% of the style-related but 58.7\% of the mathematical amplitude, lowering the ratio to \GatedStyleMathRatio{} (Figure~\ref{fig:amplitude}(d)).

\paragraph{The magnitude source sets the gated balance.}
The mask $m_t$ decides which positions and directions are retained, while the magnitude source decides their relative weight.
To isolate the magnitude source, we apply the same mask to teacher--student feedback and compare $m_tr_t$ with $m_td_t$ (Figure~\ref{fig:amplitude}(c--d)).
Gated teacher--student coefficients are smaller than raw feedback in every category, yet style-related positions still receive \GatedOPDStyleAmplitude{} versus \GatedOPDMathAmplitude{} at mathematical positions, a ratio of \GatedOPDStyleMathRatio{} compared with \AmplitudeRatio{} before filtering.
Student--base magnitudes lower the style-related amplitude by \GatedAmplitudeStyleReduction{}\% and the mathematical amplitude from \GatedOPDMathAmplitude{} to \GatedStudentMathAmplitude{}, giving a ratio of \GatedStyleMathRatio{} instead.
Because both signals retain identical positions and a uniform rescaling of OPD would leave the ratio unchanged, this difference comes from the magnitude source rather than from filtering.
These patterns also hold with another teacher and on GSM8K (Appendix~\ref{app:observation-generalization}).
These observations motivate combining student-derived strength with teacher-guided selection.

\begin{algorithm}[t]
 \caption{\SDOPD{}}
 \label{alg:sdopd}
 \small
 \begin{algorithmic}[1]
 \Require $S_0$; frozen $T,S_B$; $\mathcal{D},K,R,c_{\mathrm{OPD}}$
 \State $S_\theta\gets S_0$
 \For{$r=1,\ldots,R$}
   \State $p\gets p_{S_\theta},\quad \mathcal{R}\gets\operatorname{Rollout}(\mathcal{D},p)$
   \For{$(x,y)\in\mathcal{R},\quad t=1,\ldots,|y|$}
     \State $h_t\gets(x,y_{<t})$
     \State $\widetilde A_{S,t}\gets\approxcent_{p,K}[\log(p/p_{S_B})](y_t\mid h_t)$
     \State $\widetilde A_{\mathrm{OPD},t}\gets\approxcent_{p,K}[\log(p_T/p)](y_t\mid h_t)$
     \State $g_t\gets\ind[\widetilde A_{S,t}\widetilde A_{\mathrm{OPD},t}>0]$
     \State $\widehat A_t\gets\sg(c_{\mathrm{OPD}}g_t\widetilde A_{S,t})$
   \EndFor
   \For{each minibatch $\mathcal{B}$ from $\mathcal{R}$}
     \State $w_t\gets p_{S_\theta}(y_t\mid h_t)/p(y_t\mid h_t)$
     \State $\mathcal{L}_{\mathcal{B}}\gets\dfrac{\sum_{(x,y)\in\mathcal{B}}\sum_{t=1}^{|y|}\ell_t^{\mathrm{dual}}}{\sum_{(x,y)\in\mathcal{B}}|y|}$ \hfill (Eq.~\ref{eq:dual-clip})
     \State $\theta\gets\operatorname{AdamStep}(\theta,\nabla_\theta\mathcal{L}_{\mathcal{B}})$
   \EndFor
 \EndFor
 \State \Return $S_\theta$
 \end{algorithmic}
\end{algorithm}

\section{Method}
\label{sec:method}
Since teacher--student disagreement is larger at style-related than at mathematical positions (Section~\ref{sec:observations}), \SDOPD{} takes update strength from the student's own changes and uses the teacher to decide which directions to retain (Figure~\ref{fig:method}).

\subsection{Preliminaries: On-Policy Distillation}
\label{sec:opd-preliminaries}
For prompt $x$, write $q_\theta=p_{S_\theta}$ and $h_t=(x,y_{<t})$.
OPD samples student responses and minimizes reverse KL to a fixed teacher $T$ \citep{agarwal2024policy,gu2023minillm}:
\begin{equation}
 \begin{aligned}
  \mathcal{L}_{\mathrm{OPD}}(\theta)
  &=\mathbb{E}_{y\sim q_\theta}\!\left[\log\frac{q_\theta(y\mid x)}{p_T(y\mid x)}\right],\\
  -\nabla_\theta\mathcal{L}_{\mathrm{OPD}}
  &=\mathbb{E}_{y\sim q_\theta}\!\left[\sum_t G_t\nabla_\theta\log q_\theta(y_t\mid h_t)\right].
 \end{aligned}
 \label{eq:opd-objective}
\end{equation}
Here $G_t=\sum_{u=t}^{|y|}[\log p_T(y_u\mid h_u)-\log q_\theta(y_u\mid h_u)]$ sums log-ratios from token $t$ onward.
The score-function identity removes zero-mean terms, and KL descent reverses the log-ratio.
The zero-discount approximation keeps only the current log-ratio \citep{yang2026learning}, yielding the uncentered token advantage at rollout snapshot $p=p_{S_{\mathrm{old}}}$:
\begin{equation}
 A^{\mathrm{raw}}_{\mathrm{OPD},t}=\log p_T(y_t\mid h_t)-\log p(y_t\mid h_t).
 \label{eq:opd-feedback}
\end{equation}
Each round fixes $p$ and advantages during clipped updates, then refreshes both.
Variants share this sampling and update pipeline by replacing the advantage: ExOPD adds a scaled teacher--reference log-ratio (Appendix~\ref{app:exopd}); Our Method uses student-derived strength and teacher-selected direction.

\subsection{A Student-Derived Distillation Signal}
\label{sec:center}
We initialize the student from a post-trained model $S_0$ and freeze its pretrained base $S_B$, whose comparison with the student identifies preferences acquired through post-training (Section~\ref{sec:content}).
To measure each token's preference relative to the average shift at its prefix, we center log-ratios under $p$, with prefix conditioning implicit \citep{sutton1999policy,heo2026policy}:
\begin{equation}
 \cent_p[f](a)=f(a)-\mathbb{E}_{v\sim p}[f(v)].
 \label{eq:center}
\end{equation}
The student-derived coefficient is
\begin{equation}
 A_S=\cent_p\bigl[\log p-\log p_{S_B}\bigr].
 \label{eq:as}
\end{equation}
Positive and negative coefficients identify student--base shifts above and below the current-policy average, respectively.
Because the numerator uses the current student, each refresh incorporates preferences acquired during distillation alongside the initial post-training changes.

\subsection{Teacher-Guided Reinforcement}
Because these changes also contain stylistic preferences (Section~\ref{sec:content}), we reinforce a student-derived direction only when the teacher supports it.
Centering the raw advantage in Equation~\ref{eq:opd-feedback} gives
\begin{equation}
 A_{\mathrm{OPD}}=\cent_p\bigl[\log p_T-\log p\bigr].
 \label{eq:aopd}
\end{equation}
The distillation coefficient is
\begin{equation}
 A_{\text{SD-OPD}}
 =\underbrace{\ind\bigl[A_S A_{\mathrm{OPD}}>0\bigr]}_{g}\,A_S.
 \label{eq:sdopd}
\end{equation}
Positive agreement reinforces the sampled token, negative agreement suppresses it, and opposite signs or exact zeros yield no update.

For sampled token $y_t$, let $\widetilde A_{\text{SD-OPD},t}$ denote Equation~\ref{eq:sdopd} evaluated with approximate centering.
We use it as a fixed advantage:
\begin{equation}
 \widehat A_t=\sg\bigl(c_{\mathrm{OPD}}\,
             \widetilde A_{\text{SD-OPD},t}\bigr),
 \label{eq:advantage}
\end{equation}
where $c_{\mathrm{OPD}}$ is the distillation scale and $\sg$ stops gradients through the scores and gate.
We optimize $S_\theta$ with the dual-clipped policy loss (Appendix~\ref{app:dual-clipping}), using probability ratios relative to $p$ \citep{schulman2017proximal} and averaging over all response tokens, including zero-gated positions.
Algorithm~\ref{alg:sdopd} summarizes the complete procedure.

\paragraph{Implementation.}
Centering uses the student's top $K=1{,}024$ candidates, including the sampled token, with full-vocabulary probabilities (Appendix~\ref{app:candidate-centering}).

\begin{table*}[!t]
 \centering
 \caption{Pass@1 (\%) across \MainSettingCount{} teacher--student settings (teacher $\rightarrow$ student), averaged over 16 sampled responses per problem. Average is the six-task mean; averages and gains use unrounded scores. Gray rows at the bottom of each block give teacher reference scores; boldface marks the highest student scores within each setting, and green rows report \SDOPD{} gains over OPD in percentage points.}
 \label{tab:main}
 \small
 \setlength{\tabcolsep}{3pt}
 \begin{tabular*}{\textwidth}{@{\extracolsep{\fill}}lrrrrrrr@{}}
 \toprule
Method & AIME24 & AIME25 & AIME26 & HMMT25 & AMC23 & MATH500 & Average \\
\midrule
\multicolumn{8}{c}{\textbf{Qwen3-4B-Instruct-2507 $\boldsymbol{\rightarrow}$ Qwen3-1.7B}} \\
\addlinespace[2pt]
\mbox{Initial student} & 13.12 & 8.54 & 7.71 & 6.67 & 44.69 & 69.40 & 25.02 \\
OPD & 32.08 & 26.25 & 24.38 & 16.88 & 73.75 & 83.90 & 42.87 \\
ExOPD & 40.42 & 28.12 & 28.75 & 20.00 & 79.06 & 86.11 & 47.08 \\
\SDOPD{} & \textbf{45.42} & \textbf{33.33} & \textbf{40.62} & \textbf{21.04} & \textbf{86.72} & \textbf{88.72} & \textbf{52.64} \\
\textcolor{improvementgreen}{$\Delta$ vs. OPD} & \textcolor{improvementgreen}{+13.33} & \textcolor{improvementgreen}{+7.08} & \textcolor{improvementgreen}{+16.25} & \textcolor{improvementgreen}{+4.17} & \textcolor{improvementgreen}{+12.97} & \textcolor{improvementgreen}{+4.83} & \textcolor{improvementgreen}{+9.77} \\
\textcolor{teachergray}{Teacher} & \textcolor{teachergray}{58.75} & \textcolor{teachergray}{44.38} & \textcolor{teachergray}{53.12} & \textcolor{teachergray}{29.17} & \textcolor{teachergray}{93.44} & \textcolor{teachergray}{93.33} & \textcolor{teachergray}{62.03} \\
\midrule
\multicolumn{8}{c}{\textbf{Qwen3-30B-A3B-Instruct-2507 $\boldsymbol{\rightarrow}$ Qwen3-1.7B}} \\
\addlinespace[2pt]
\mbox{Initial student} & 13.12 & 8.54 & 7.71 & 6.67 & 44.69 & 69.40 & 25.02 \\
OPD & 31.87 & 27.08 & 26.67 & 16.67 & 72.03 & 84.60 & 43.15 \\
ExOPD & 35.00 & 29.38 & 30.83 & 17.71 & 71.09 & 85.84 & 44.97 \\
\SDOPD{} & \textbf{44.79} & \textbf{37.08} & \textbf{40.21} & \textbf{18.33} & \textbf{88.91} & \textbf{89.53} & \textbf{53.14} \\
\textcolor{improvementgreen}{$\Delta$ vs. OPD} & \textcolor{improvementgreen}{+12.92} & \textcolor{improvementgreen}{+10.00} & \textcolor{improvementgreen}{+13.54} & \textcolor{improvementgreen}{+1.67} & \textcolor{improvementgreen}{+16.88} & \textcolor{improvementgreen}{+4.93} & \textcolor{improvementgreen}{+9.99} \\
\textcolor{teachergray}{Teacher} & \textcolor{teachergray}{77.08} & \textcolor{teachergray}{60.00} & \textcolor{teachergray}{72.29} & \textcolor{teachergray}{42.50} & \textcolor{teachergray}{95.31} & \textcolor{teachergray}{95.74} & \textcolor{teachergray}{73.82} \\
\midrule
\multicolumn{8}{c}{\textbf{MiMo-7B-RL-0530 $\boldsymbol{\rightarrow}$ Qwen3-1.7B}} \\
\addlinespace[2pt]
\mbox{Initial student} & 13.12 & 8.54 & 7.71 & 6.67 & 44.69 & 69.40 & 25.02 \\
OPD & 43.54 & 36.67 & 34.58 & 21.25 & 79.84 & 88.01 & 50.65 \\
ExOPD & 35.42 & 30.00 & 27.50 & 20.62 & 67.81 & 85.02 & 44.40 \\
\SDOPD{} & \textbf{54.37} & \textbf{43.33} & \textbf{45.21} & \textbf{31.04} & \textbf{90.94} & \textbf{90.84} & \textbf{59.29} \\
\textcolor{improvementgreen}{$\Delta$ vs. OPD} & \textcolor{improvementgreen}{+10.83} & \textcolor{improvementgreen}{+6.67} & \textcolor{improvementgreen}{+10.63} & \textcolor{improvementgreen}{+9.79} & \textcolor{improvementgreen}{+11.09} & \textcolor{improvementgreen}{+2.83} & \textcolor{improvementgreen}{+8.64} \\
\textcolor{teachergray}{Teacher} & \textcolor{teachergray}{75.21} & \textcolor{teachergray}{63.12} & \textcolor{teachergray}{63.75} & \textcolor{teachergray}{42.92} & \textcolor{teachergray}{86.56} & \textcolor{teachergray}{89.61} & \textcolor{teachergray}{70.20} \\
\midrule
\multicolumn{8}{c}{\textbf{Qwen3-30B-A3B-Instruct-2507 $\boldsymbol{\rightarrow}$ Qwen3-4B}} \\
\addlinespace[2pt]
\mbox{Initial student} & 21.46 & 18.54 & 14.37 & 11.67 & 68.91 & 81.49 & 36.07 \\
OPD & 60.62 & 46.25 & 54.79 & 33.54 & 89.22 & 92.34 & 62.79 \\
ExOPD & 66.67 & 55.21 & 60.00 & 35.42 & 92.03 & \textbf{93.27} & 67.10 \\
\SDOPD{} & \textbf{69.79} & \textbf{62.29} & \textbf{60.62} & \textbf{42.08} & \textbf{95.78} & 92.74 & \textbf{70.55} \\
\textcolor{improvementgreen}{$\Delta$ vs. OPD} & \textcolor{improvementgreen}{+9.17} & \textcolor{improvementgreen}{+16.04} & \textcolor{improvementgreen}{+5.83} & \textcolor{improvementgreen}{+8.54} & \textcolor{improvementgreen}{+6.56} & \textcolor{improvementgreen}{+0.40} & \textcolor{improvementgreen}{+7.76} \\
\textcolor{teachergray}{Teacher} & \textcolor{teachergray}{77.08} & \textcolor{teachergray}{60.00} & \textcolor{teachergray}{72.29} & \textcolor{teachergray}{42.50} & \textcolor{teachergray}{95.31} & \textcolor{teachergray}{95.74} & \textcolor{teachergray}{73.82} \\
\midrule
\multicolumn{8}{c}{\textbf{Qwen3-30B-A3B-Instruct-2507 $\boldsymbol{\rightarrow}$ Qwen3-8B}} \\
\addlinespace[2pt]
\mbox{Initial student} & 26.46 & 20.42 & 15.42 & 10.83 & 66.72 & 82.35 & 37.03 \\
OPD & 67.71 & 54.37 & 55.00 & 32.71 & 91.56 & 93.11 & 65.74 \\
ExOPD & 68.96 & 51.25 & 61.46 & 32.92 & 93.91 & \textbf{93.75} & 67.04 \\
\SDOPD{} & \textbf{80.42} & \textbf{68.96} & \textbf{67.50} & \textbf{48.75} & \textbf{97.34} & 93.70 & \textbf{76.11} \\
\textcolor{improvementgreen}{$\Delta$ vs. OPD} & \textcolor{improvementgreen}{+12.71} & \textcolor{improvementgreen}{+14.58} & \textcolor{improvementgreen}{+12.50} & \textcolor{improvementgreen}{+16.04} & \textcolor{improvementgreen}{+5.78} & \textcolor{improvementgreen}{+0.59} & \textcolor{improvementgreen}{+10.37} \\
\textcolor{teachergray}{Teacher} & \textcolor{teachergray}{77.08} & \textcolor{teachergray}{60.00} & \textcolor{teachergray}{72.29} & \textcolor{teachergray}{42.50} & \textcolor{teachergray}{95.31} & \textcolor{teachergray}{95.74} & \textcolor{teachergray}{73.82} \\
 
 \bottomrule
 \end{tabular*}
\end{table*}

\section{Experiments}
\label{sec:experiments}
\subsection{Experimental Setup}
\label{sec:setup}
\paragraph{Models.}
We evaluate \MainSettingCount{} teacher--student settings with Qwen3 students \citep{yang2025qwen3} of 1.7B, 4B, and 8B parameters (Table~\ref{tab:main}).
The teachers are Qwen3-4B-Instruct-2507, Qwen3-30B-A3B-Instruct-2507, and MiMo-7B-RL-0530; each table block identifies the teacher and student for that setting.
\SDOPD{} uses each student's corresponding pretrained base checkpoint as the frozen reference.
Implementation details for the MiMo teacher are provided in Appendix~\ref{app:mimo-compatibility}.
We compare \SDOPD{} with the initial student, OPD, and ExOPD \citep{yang2026learning}, whose formulation is detailed in Appendix~\ref{app:exopd}.

\begin{table*}[!t]
 \centering
 \caption{Teacher guidance and source refresh ablations for the 1.7B student, using matched training and evaluation settings. \Frozen{} substitutes the initial student in the student--base difference. \Ungated{} removes the teacher sign gate and teacher scoring.}
 \label{tab:ablations}
 \small
 \setlength{\tabcolsep}{4pt}
 \begin{tabular}{@{}lrrrrrrr@{}}
 \toprule
 Variant & AIME24 & AIME25 & AIME26 & HMMT25 & AMC23 & MATH500 & Average \\
 \midrule
\SDOPD{} & \textbf{45.42} & \textbf{33.33} & \textbf{40.62} & \textbf{21.04} & \textbf{86.72} & \textbf{88.72} & \textbf{52.64} \\
Frozen & 23.75 & 17.08 & 13.12 & 11.25 & 63.12 & 78.46 & 34.47 \\
Ungated & 0.00 & 0.00 & 0.00 & 0.00 & 0.00 & 0.00 & 0.00 \\
 
 \bottomrule
 \end{tabular}
\end{table*}

\paragraph{Training.}
All trained models use DAPO-Math-17K prompts \citep{yu2026dapo} for 272 rollout rounds, corresponding to one full pass over the training dataset (one epoch).
Each round samples 64 prompts and four responses per prompt.
The learning rate is $10^{-6}$ and the response batch size is 64.
We match rollout and response budgets across methods and evaluate their final checkpoints.
Appendices~\ref{app:training} and~\ref{app:dual-clipping} detail the shared training hyperparameters and method-specific clipping settings, respectively.

\paragraph{Evaluation.}
We evaluate on six mathematics benchmarks: AIME24 and AIME25 \citep{zhuzilin2025aime24,zhuzilin2025aime25}, AIME26 and HMMT25 (February) from MathArena \citep{dekoninck2026beyond}, AMC23 \citep{mathai2025amc23}, and MATH500 \citep{hendrycks2021measuring,lightman2024let}.
We sample 16 responses per problem in non-thinking mode and report pass@1 averaged over these samples (avg@16) on each benchmark.
The evaluation protocol uses temperature 0.7, top-$p$ 0.8, top-$k$ 20, and a maximum response length of 38,912 tokens.
We retain completed and truncated responses and check extracted final answers for normalized or symbolic equivalence.

\subsection{Main Results}
\paragraph{Gains over OPD on every benchmark and setting.}
Table~\ref{tab:main} shows that \SDOPD{} improves six-task average pass@1 over OPD by \MinOPDSettingGain{}--\MaxOPDSettingGain{} percentage points across all \MainSettingCount{} teacher--student configurations.
The improvements hold on every benchmark in every configuration, covering all \MainComparisonCount{} comparisons under matched rollout and response budgets, and \SDOPD{} achieves the highest six-task student average in every configuration.

\paragraph{Gains across teachers.}
The gains do not depend on a particular teacher family.
For Qwen3-1.7B, average gains over OPD are \MainGain{}, \LargeTeacherSmallStudentGain{}, and \MiMoGain{} percentage points for Qwen3-4B-Instruct-2507, Qwen3-30B-A3B-Instruct-2507, and MiMo-7B-RL-0530 teachers, respectively.
ExOPD improves the average with both Qwen3 teachers, but falls below OPD with MiMo: \MiMoExOPDAverage{}\% versus \MiMoOPDAverage{}\%, a \MiMoExOPDDrop{}-point decrease.
In the same MiMo setting, \SDOPD{} reaches \MiMoSDOPDAverage{}\%, exceeding ExOPD by \MiMoExOPDGap{} points under the shared-vocabulary protocol in Appendix~\ref{app:mimo-compatibility}.

\paragraph{Gains across student sizes.}
Using Qwen3-30B-A3B-Instruct-2507 as the teacher, average gains over OPD are \LargeTeacherSmallStudentGain{}, \FourBGain{}, and \EightBGain{} percentage points for the 1.7B, 4B, and 8B students, respectively.
The margin over ExOPD is smallest for the 4B student (\FourBExOPDGap{} points) and widens to \EightBExOPDGap{} points for the 8B student.
Across all \MainSettingCount{} configurations, \SDOPD{} also exceeds ExOPD on \ExOPDBenchmarkWins{} of the \MainComparisonCount{} benchmark comparisons.

\paragraph{Students can surpass their teachers.}
Distilled from Qwen3-30B-A3B-Instruct-2507, the 8B student reaches \EightBSDOPDAverage{}\% average pass@1, exceeding the teacher's \LargeTeacherAverage{}\% by \EightBTeacherGain{} points, while OPD and ExOPD remain below it; this student also exceeds the teacher on AIME24, AIME25, HMMT25, and AMC23.
On individual benchmarks, the 4B student also exceeds this teacher on AIME25 and AMC23, and the 1.7B student exceeds MiMo on AMC23 and MATH500.

\subsection{Ablation Study}
\label{sec:ablations}
The two ablations test the two design choices of Section~\ref{sec:method} in the 4B-to-1.7B setting: \Ungated{} asks whether student-derived strength needs teacher selection, and \Frozen{} asks whether the student signal must be refreshed.

\paragraph{Teacher guidance.}
\Ungated{} uses $A_S$ directly, with the same candidate approximation, replacing $gA_S$ and removing teacher scoring.
Its pass@1 is zero on all six tasks (Table~\ref{tab:ablations}).
All \EvaluationResponseCount{} responses complete within \UngatedMinLength{}--\UngatedMaxLength{} tokens, including the termination token, with a mean of \UngatedMeanLength{} tokens.
Appendix~\ref{app:failure} traces how these short, low-entropy responses emerge and explains them with a local gradient analysis: without the gate, the student--base signal reinforces deviations from the base.

\paragraph{Student-source refresh.}
\Frozen{} replaces the current student in the numerator of the student--base ratio with the fixed initial student:
\begin{equation}
 A_S^{(0)}=\cent_p[\log p_{S_0}-\log p_{S_B}].
 \label{eq:frozen-student}
\end{equation}
\begin{equation}
 A_{\mathrm{Frozen}}
 =\underbrace{\ind\bigl[A_S^{(0)}A_{\mathrm{OPD}}>0\bigr]}_{g_0}
 \,A_S^{(0)}.
 \label{eq:frozen}
\end{equation}
\Frozen{} uses the candidate approximation in Equation~\ref{eq:approx}; only the student--base source is fixed at $S_0$, while candidate selection, centering weights, rollout generation, and teacher feedback follow the current policy $p$.

\Frozen{} attains \FrozenAverage{}\% average pass@1, \FrozenGap{} points below \SDOPD{} and also below OPD (\OPDAverage{}\%).
Without refresh, both the coefficient magnitudes and the gate decisions follow the preferences of $S_0$, so supervision cannot build on what the student acquires during distillation.

\subsection{Training Dynamics}
\label{sec:training-dynamics}
\begin{figure}[t]
 \centering
 \includegraphics[width=\linewidth]{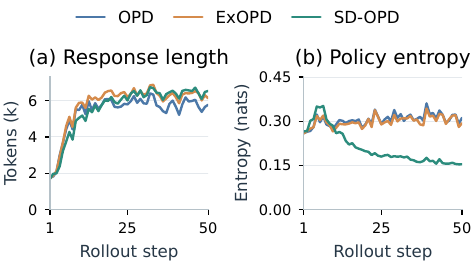}
 \caption{Early training dynamics with Qwen3-30B-A3B-Instruct-2507 teaching Qwen3-1.7B. Each step is one rollout round with four optimizer updates. Length is the rollout mean; entropy averages the four logged update means. Curves are unsmoothed. Appendix~\ref{app:sdopd-dynamics} shows all 272 training steps.}
 \label{fig:training-dynamics}
\end{figure}

To see how the choice of coefficient shapes the policy during training, Figure~\ref{fig:training-dynamics} tracks response length and policy entropy for OPD, ExOPD, and \SDOPD{} in the 30B-to-1.7B setting.
We focus on the first 50 rollout steps, where the main differences emerge.

\paragraph{All three methods rapidly lengthen responses.}
Mean response length rises from approximately 1.7k tokens to around 6k within the first 20 steps.
Over steps 41--50, OPD, ExOPD, and \SDOPD{} average 5.76k, 6.28k, and 6.45k tokens, respectively.
ExOPD and \SDOPD{} reach similar lengths in this phase, although they build their coefficients differently.

\paragraph{\SDOPD{} develops a more concentrated policy.}
The entropy curves distinguish the methods more clearly.
After an initial rise, \SDOPD{} entropy declines to an average of 0.158 nats over steps 41--50, compared with 0.314 for OPD and 0.308 for ExOPD.
\SDOPD{} therefore keeps responses long while concentrating next-token predictions at the prefixes it visits.
This pattern is consistent with reinforcing selected student preferences.
Beyond step 50, the separation persists, with further gradual entropy reduction in \SDOPD{} and continuing length fluctuations (Appendix~\ref{app:sdopd-dynamics}).

\section{Conclusion}
We showed that a student's existing post-training changes can drive continued on-policy distillation.
Since teacher--student disagreement is several times larger at style-related than at mathematical positions, \SDOPD{} instead takes update strength from student--base differences, constrains their direction with teacher feedback, and refreshes the student source throughout training.
Across \MainSettingCount{} teacher--student configurations, \SDOPD{} improves all six mathematics benchmarks over OPD and achieves higher six-task averages than both OPD and ExOPD.
Average gains over OPD range from \MinOPDSettingGain{} to \MaxOPDSettingGain{} percentage points when either the teacher or the student size changes.

The ablations show that both components are needed: reinforcing student preferences without teacher feedback collapses all six benchmarks to zero, and fixing the student source falls below OPD.
Further work can examine which of the student's own post-training changes provide useful supervision, how their value evolves during training, and how they can support continued learning across tasks.

\section*{Limitations}
Our evaluation covers three teacher checkpoints from two families (Qwen3 and MiMo) and Qwen3 students of 1.7B, 4B, and 8B parameters on mathematical reasoning; extending \SDOPD{} to larger students and other domains is a natural next step.
The comparisons match rollout and response budgets (clipping settings in Appendix~\ref{app:dual-clipping}), and the token analysis uses fixed rule-based categories that apply unchanged across models and datasets.
\SDOPD{} uses the student's pretrained base, released with open model families such as Qwen3, and adds one base scoring pass per round.
Top-$K$ centering ($K=1{,}024$) evaluates the centering sums over a shared candidate set at each prefix, using full-vocabulary probabilities.

\bibliography{custom,references_pending}

@inproceedings{agarwal2024policy,
  title={On-policy distillation of language models: Learning from self-generated mistakes},
  author={Agarwal, Rishabh and Vieillard, Nino and Zhou, Yongchao and Stanczyk, Piotr and Ramos Garea, Sabela and Geist, Matthieu and Bachem, Olivier},
  booktitle={International Conference on Learning Representations},
  volume={2024},
  pages={21246--21263},
  year={2024}
}

@article{gu2023minillm,
  title={Minillm: On-policy distillation of large language models},
  author={Gu, Yuxian and Dong, Li and Wei, Furu and Huang, Minlie},
  journal={arXiv preprint arXiv:2306.08543},
  year={2023}
}

@inproceedings{lin2024unlocking,
  title={The unlocking spell on base llms: Rethinking alignment via in-context learning},
  author={Lin, Bill Yuchen and Ravichander, Abhilasha and Lu, Ximing and Dziri, Nouha and Sclar, Melanie and Chandu, Khyathi and Bhagavatula, Chandra and Choi, Yejin},
  booktitle={International Conference on Learning Representations},
  volume={2024},
  pages={24907--24933},
  year={2024}
}

@article{fu2026revisiting,
  title={Revisiting on-policy distillation: Empirical failure modes and simple fixes},
  author={Fu, Yuqian and Huang, Haohuan and Jiang, Kaiwen and Liu, Jiacai and Jiang, Zhuo and Zhu, Yuanheng and Zhao, Dongbin},
  journal={arXiv preprint arXiv:2603.25562},
  year={2026}
}

@article{wang2026dr,
  title={Dr. OPD: Learning What to Follow for Optimal On-Policy Distillation of Large Language Models},
  author={Wang, Zhenyu and Wang, Tianze and Zhang, Linjun and Hu, Yifan},
  journal={arXiv preprint arXiv:2609.38025},
  year={2026}
}

@article{zhao2026self,
  title={Self-distilled reasoner: On-policy self-distillation for large language models},
  author={Zhao, Siyan and Xie, Zhihui and Liu, Mengchen and Huang, Jing and Pang, Guan and Chen, Feiyu and Grover, Aditya},
  journal={arXiv preprint arXiv:2601.18734},
  year={2026}
}

@article{ye2026policy,
  title={On-policy context distillation for language models},
  author={Ye, Tianzhu and Dong, Li and Wu, Xun and Huang, Shaohan and Wei, Furu},
  journal={arXiv preprint arXiv:2602.12275},
  year={2026}
}

@article{yang2026learning,
  title={Learning beyond teacher: Generalized on-policy distillation with reward extrapolation},
  author={Yang, Wenkai and Liu, Weijie and Xie, Ruobing and Yang, Kai and Yang, Saiyong and Lin, Yankai},
  journal={arXiv preprint arXiv:2602.12125},
  year={2026}
}

@article{heo2026policy,
  title={On-Policy Delta Distillation},
  author={Heo, Byeongho and Hwang, Jaehui and Yun, Sangdoo and Han, Dongyoon},
  journal={arXiv preprint arXiv:2607.15161},
  year={2026}
}

@article{yu2026dapo,
  title={Dapo: An open-source llm reinforcement learning system at scale},
  author={Yu, Qiying and Zhang, Zheng and Zhu, Ruofei and Yuan, Yufeng and Zuo, Xiaochen and Yue, Yu and Dai, Weinan and Fan, Tiantian and Liu, Gaohong and Liu, Lingjun and others},
  journal={Advances in Neural Information Processing Systems},
  volume={38},
  pages={113222--113244},
  year={2026}
}

@article{yang2025qwen3,
  title={Qwen3 technical report},
  author={Yang, An and Li, Anfeng and Yang, Baosong and Zhang, Beichen and Hui, Binyuan and Zheng, Bo and Yu, Bowen and Gao, Chang and Huang, Chengen and Lv, Chenxu and others},
  journal={arXiv preprint arXiv:2505.09388},
  year={2025}
}

@article{cobbe2021training,
  title={Training verifiers to solve math word problems},
  author={Cobbe, Karl and Kosaraju, Vineet and Bavarian, Mohammad and Chen, Mark and Jun, Heewoo and Kaiser, Lukasz and Plappert, Matthias and Tworek, Jerry and Hilton, Jacob and Nakano, Reiichiro and others},
  journal={arXiv preprint arXiv:2110.14168},
  year={2021}
}

@article{schulman2017proximal,
  title={Proximal policy optimization algorithms},
  author={Schulman, John and Wolski, Filip and Dhariwal, Prafulla and Radford, Alec and Klimov, Oleg},
  journal={arXiv preprint arXiv:1707.06347},
  year={2017}
}

@inproceedings{ye2020mastering,
  title={Mastering complex control in moba games with deep reinforcement learning},
  author={Ye, Deheng and Liu, Zhao and Sun, Mingfei and Shi, Bei and Zhao, Peilin and Wu, Hao and Yu, Hongsheng and Yang, Shaojie and Wu, Xipeng and Guo, Qingwei and others},
  booktitle={Proceedings of the AAAI conference on artificial intelligence},
  volume={34},
  number={04},
  pages={6672--6679},
  year={2020}
}

@article{hinton2015distilling,
  title={Distilling the knowledge in a neural network},
  author={Hinton, Geoffrey and Vinyals, Oriol and Dean, Jeff},
  journal={arXiv preprint arXiv:1503.02531},
  year={2015}
}

@inproceedings{kim2016sequence,
  title={Sequence-level knowledge distillation},
  author={Kim, Yoon and Rush, Alexander M},
  booktitle={Proceedings of the 2016 conference on empirical methods in natural language processing},
  pages={1317--1327},
  year={2016}
}

@article{xiao2026mimo,
  title={Mimo-v2-flash technical report},
  author={Xiao, Bangjun and Xia, Bingquan and Yang, Bo and Gao, Bofei and Shen, Bowen and Zhang, Chen and He, Chenhong and Lou, Chiheng and Luo, Fuli and Wang, Gang and others},
  journal={arXiv preprint arXiv:2601.02780},
  year={2026}
}

@article{rafailov2023direct,
  title={Direct preference optimization: Your language model is secretly a reward model},
  author={Rafailov, Rafael and Sharma, Archit and Mitchell, Eric and Manning, Christopher D and Ermon, Stefano and Finn, Chelsea},
  journal={Advances in neural information processing systems},
  volume={36},
  pages={53728--53741},
  year={2023}
}

@article{sutton1999policy,
  title={Policy gradient methods for reinforcement learning with function approximation},
  author={Sutton, Richard S and McAllester, David and Singh, Satinder and Mansour, Yishay},
  journal={Advances in neural information processing systems},
  volume={12},
  year={1999}
}

@inproceedings{kwon2023efficient,
  title={Efficient memory management for large language model serving with pagedattention},
  author={Kwon, Woosuk and Li, Zhuohan and Zhuang, Siyuan and Sheng, Ying and Zheng, Lianmin and Yu, Cody Hao and Gonzalez, Joseph and Zhang, Hao and Stoica, Ion},
  booktitle={Proceedings of the 29th symposium on operating systems principles},
  pages={611--626},
  year={2023}
}

@inproceedings{wolf2020transformers,
  title={Transformers: State-of-the-art natural language processing},
  author={Wolf, Thomas and Debut, Lysandre and Sanh, Victor and Chaumond, Julien and Delangue, Clement and Moi, Anthony and Cistac, Pierric and Rault, Tim and Louf, R{\'e}mi and Funtowicz, Morgan and others},
  booktitle={Proceedings of the 2020 conference on empirical methods in natural language processing: system demonstrations},
  pages={38--45},
  year={2020}
}

@inproceedings{lightman2024let,
  title={Let's verify step by step},
  author={Lightman, Hunter and Kosaraju, Vineet and Burda, Yuri and Edwards, Harrison and Baker, Bowen and Lee, Teddy and Leike, Jan and Schulman, John and Sutskever, Ilya and Cobbe, Karl},
  booktitle={International Conference on Learning Representations},
  volume={2024},
  pages={39578--39601},
  year={2024}
}

@article{hendrycks2021measuring,
  title={Measuring mathematical problem solving with the math dataset},
  author={Hendrycks, Dan and Burns, Collin and Kadavath, Saurav and Arora, Akul and Basart, Steven and Tang, Eric and Song, Dawn and Steinhardt, Jacob},
  journal={arXiv preprint arXiv:2103.03874},
  year={2021}
}

@article{dekoninck2026beyond,
  title={Beyond benchmarks: Matharena as an evaluation platform for mathematics with llms},
  author={Dekoninck, Jasper and Jovanovi{\'c}, Nikola and Gehrunger, Tim and R{\"o}gnvaldsson, K{\'a}ri and Petrov, Ivo and Sun, Chenhao and Vechev, Martin},
  journal={arXiv preprint arXiv:2605.00674},
  year={2026}
}

@misc{zhuzilin2025aime24,
  title = {{AIME 2024}},
  author = {{zhuzilin}},
  howpublished = {Hugging Face dataset},
  year = {2025},
  url = {https://huggingface.co/datasets/zhuzilin/aime-2024}
}

@misc{zhuzilin2025aime25,
  title = {{AIME 2025}},
  author = {{zhuzilin}},
  howpublished = {Hugging Face dataset},
  year = {2025},
  url = {https://huggingface.co/datasets/zhuzilin/aime-2025}
}

@misc{mathai2025amc23,
  title = {{AMC23}},
  author = {{math-ai}},
  howpublished = {Hugging Face dataset},
  year = {2025},
  url = {https://huggingface.co/datasets/math-ai/amc23}
}

@article{ko2024distillm,
  title={Distillm: Towards streamlined distillation for large language models},
  author={Ko, Jongwoo and Kim, Sungnyun and Chen, Tianyi and Yun, Se-Young},
  journal={arXiv preprint arXiv:2402.03898},
  year={2024}
}

@article{ko2025distillm,
  title={Distillm-2: A contrastive approach boosts the distillation of llms},
  author={Ko, Jongwoo and Chen, Tianyi and Kim, Sungnyun and Ding, Tianyu and Liang, Luming and Zharkov, Ilya and Yun, Se-Young},
  journal={arXiv preprint arXiv:2503.07067},
  year={2025}
}

@article{ye2025black,
  title={Black-box on-policy distillation of large language models},
  author={Ye, Tianzhu and Dong, Li and Chi, Zewen and Wu, Xun and Huang, Shaohan and Wei, Furu},
  journal={arXiv preprint arXiv:2511.10643},
  year={2025}
}

@article{hubotter2026reinforcement,
  title={Reinforcement learning via self-distillation},
  author={H{\"u}botter, Jonas and L{\"u}beck, Frederike and Behric, Lejs and Baumann, Anton and Bagatella, Marco and Marta, Daniel and Hakimi, Ido and Shenfeld, Idan and Buening, Thomas Kleine and Guestrin, Carlos and others},
  journal={arXiv preprint arXiv:2601.20802},
  year={2026}
}

@article{shenfeld2026self,
  title={Self-distillation enables continual learning},
  author={Shenfeld, Idan and Damani, Mehul and H{\"u}botter, Jonas and Agrawal, Pulkit},
  journal={arXiv preprint arXiv:2601.19897},
  year={2026}
}

@article{yuan2024free,
  title={Free process rewards without process labels},
  author={Yuan, Lifan and Li, Wendi and Chen, Huayu and Cui, Ganqu and Ding, Ning and Zhang, Kaiyan and Zhou, Bowen and Liu, Zhiyuan and Peng, Hao},
  journal={arXiv preprint arXiv:2412.01981},
  year={2024}
}

@article{zhou2023lima,
  title={Lima: Less is more for alignment},
  author={Zhou, Chunting and Liu, Pengfei and Xu, Puxin and Iyer, Srinivasan and Sun, Jiao and Mao, Yuning and Ma, Xuezhe and Efrat, Avia and Yu, Ping and Yu, Lili and others},
  journal={Advances in Neural Information Processing Systems},
  volume={36},
  pages={55006--55021},
  year={2023}
}

@article{guo2025deepseek,
  title={Deepseek-r1: Incentivizing reasoning capability in llms via reinforcement learning},
  author={Guo, Daya and Yang, Dejian and Zhang, Haowei and Song, Junxiao and Wang, Peiyi and Zhu, Qihao and Xu, Runxin and Zhang, Ruoyu and Ma, Shirong and Bi, Xiao and others},
  journal={arXiv preprint arXiv:2501.12948},
  year={2025}
}

@inproceedings{ross2011reduction,
  title={A reduction of imitation learning and structured prediction to no-regret online learning},
  author={Ross, St{\'e}phane and Gordon, Geoffrey and Bagnell, Drew},
  booktitle={Proceedings of the fourteenth international conference on artificial intelligence and statistics},
  pages={627--635},
  year={2011},
  organization={JMLR Workshop and Conference Proceedings}
}

@inproceedings{guha2026openthoughts,
  title={Openthoughts: Data recipes for reasoning models},
  author={Guha, Etash and Marten, Ryan and Keh, Sedrick and Raoof, Negin and Smyrnis, Georgios and Bansal, Hritik and Nezhurina, Marianna and Mercat, Jean and Vu, Trung and Sprague, Zayne and others},
  booktitle={International Conference on Learning Representations},
  volume={2026},
  pages={108059--108130},
  year={2026}
}

@article{cui2025process,
  title={Process reinforcement through implicit rewards},
  author={Cui, Ganqu and Yuan, Lifan and Wang, Zefan and Wang, Hanbin and Zhang, Yuchen and Chen, Jiacheng and Li, Wendi and He, Bingxiang and Fan, Yuchen and Yu, Tianyu and others},
  journal={arXiv preprint arXiv:2502.01456},
  year={2025}
}

@article{zheng2024sglang,
  title={Sglang: Efficient execution of structured language model programs},
  author={Zheng, Lianmin and Yin, Liangsheng and Xie, Zhiqiang and Sun, Chuyue and Huang, Jeff and Yu, Cody H and Cao, Shiyi and Kozyrakis, Christos and Stoica, Ion and Gonzalez, Joseph E and others},
  journal={Advances in neural information processing systems},
  volume={37},
  pages={62557--62583},
  year={2024}
}
\bibliographystyle{icml2026}
\clearpage
\appendix
\section{Detailed Training Settings}
\label{app:training-settings}
\setcounter{table}{0}
\renewcommand{\thetable}{\thesection\arabic{table}}
\setcounter{figure}{0}
\renewcommand{\thefigure}{\thesection\arabic{figure}}

\subsection{Training Hyperparameters}
\label{app:training}
Table~\ref{tab:train-config} lists the training hyperparameters shared by OPD and \SDOPD{} across the reported student sizes.
Each round samples 64 prompts and four responses per prompt, yielding 256 responses for four optimizer updates with a global batch size of 64.
Training comprises 272 rounds and 1,088 updates, and the rollout policy is synchronized after every round.
We use the final checkpoint for evaluation.

\begin{table}[!htbp]
 \centering
 \caption{Shared training hyperparameters for OPD and \SDOPD{}. The same settings are used for the \Frozen{} and \Ungated{} ablations.}
 \label{tab:train-config}
 \small
 \setlength{\tabcolsep}{3pt}
 \begin{tabular}{@{}lc@{}}
 \toprule
 Hyperparameter & Value \\
 \midrule
 Rollout rounds & 272 \\
 Prompts per round & 64 \\
 Responses per prompt & 4 \\
 Response batch / microbatch & 64 / 1 \\
 Updates per round / total & 4 / 1,088 \\
 Maximum prompt tokens & 1,024 \\
 Maximum response tokens & 7,168 \\
 Maximum context tokens & 8,192 \\
 Rollout temperature & 1.0 \\
 Rollout top-$p$ & 1.0 \\
 Rollout top-$k$ & Disabled \\
 Learning rate & $10^{-6}$ \\
 Learning-rate schedule & Constant \\
 Warmup updates & 0 \\
 Optimizer & Adam \\
 Weight decay & 0.1 \\
 Adam $(\beta_1,\beta_2)$ & $(0.9,0.98)$ \\
 Adam $\epsilon$ & $10^{-8}$ \\
 Gradient norm clipping & 1.0 \\
 Precision & BF16 \\
 PPO lower / upper clipping & 0.2 / 0.2 \\
 Distillation coefficient & 1.0 \\
 Advantage normalization & Disabled \\
 Entropy coefficient & 0 \\
 Auxiliary reference KL & Disabled \\
 Training / rollout seed & 11 / 11 \\
 \bottomrule
 \end{tabular}
\end{table}

Both methods use the non-thinking template and optimize the policy loss over response tokens, with no task reward or auxiliary reference-KL penalty.
For Qwen3 teachers, OPD obtains sampled-token teacher scores through SGLang \citep{zheng2024sglang}, whereas \SDOPD{} obtains teacher and base scores over the shared candidate set defined in Appendix~\ref{app:candidate-centering}.
The candidate count $K=1{,}024$ controls the centering approximation and is independent of rollout top-$k$ sampling.
ExOPD uses the corresponding OPD training configuration; Appendix~\ref{app:exopd} specifies its update rule and extrapolation scale.

\subsection{Candidate-Set Centering}
\label{app:candidate-centering}
We approximate the expectation in Equation~\ref{eq:center} using a shared candidate set $V_K(h,y_t)$ at each prefix $h$.
The set contains the current rollout policy's top $K=1{,}024$ tokens; if the sampled token $y_t$ is absent, it replaces the lowest-probability candidate.
All three models score this same set using full-vocabulary probabilities, without renormalization within the set:
\begin{equation}
 \begin{split}
  \approxcent_{p,K}[f](y_t\mid h)
  &= f(y_t\mid h)\\
  &\quad-\sum_{v\in V_K}p(v\mid h)f(v\mid h).
 \end{split}
 \label{eq:approx}
\end{equation}
Applying this operator to the student--base and teacher--student log-ratios gives $\widetilde A_{S,t}$ and $\widetilde A_{\mathrm{OPD},t}$, respectively.
Their signs determine the gate, and their gated student-derived coefficient supplies the fixed advantage in Equation~\ref{eq:advantage}.
The candidate set, centering weights, and coefficients are recomputed from each new rollout-policy snapshot and remain fixed during its associated updates.

\subsection{Dual Clipping}
\label{app:dual-clipping}
We use the PPO clipped objective with $\epsilon=0.2$ \citep{schulman2017proximal}.
For a response token $y_t$ at prefix $h_t$, let $w_t=p_{S_\theta}(y_t\mid h_t)/p(y_t\mid h_t)$ be the policy ratio relative to the rollout policy.
Using the fixed advantage $\widehat A_t$ from Equation~\ref{eq:advantage}, the token loss is
\begin{equation}
 \begin{split}
  \ell_t^{\mathrm{PPO}}=\max\bigl\{
  &-w_t\widehat A_t,\\
  &-\operatorname{clip}(w_t,1-\epsilon,1+\epsilon)
       \widehat A_t\bigr\}.
 \end{split}
 \label{eq:ppo-clip}
\end{equation}
For a negative advantage, the first branch grows as $w_t|\widehat A_t|$ when $w_t$ increases.
Ordinary PPO clipping therefore does not bound the loss from large probability ratios on negatively weighted tokens.
\SDOPD{} is exposed to this case because teacher-supported negative student--base coefficients actively suppress tokens, and the optimized policy can drift from the fixed rollout policy across minibatch updates; a few large ratios can then dominate the update.

Dual clipping adds an upper bound to this negative-advantage branch \citep{ye2020mastering}:
\begin{equation}
 \ell_t^{\mathrm{dual}}=
 \begin{cases}
  \min\{\ell_t^{\mathrm{PPO}},-c_{\mathrm{dual}}\widehat A_t\},
       &\widehat A_t<0,\\
  \ell_t^{\mathrm{PPO}}, &\widehat A_t\geq 0.
 \end{cases}
 \label{eq:dual-clip}
\end{equation}
We set $c_{\mathrm{dual}}=3.0$ for \SDOPD{}, \Frozen{}, and \Ungated{}; the OPD and ExOPD baselines with Qwen3 teachers use ordinary PPO clipping.
The additional bound caps the negative-advantage loss at $3|\widehat A_t|$, while positive advantages retain the standard PPO objective.
The loss is averaged over all response-token positions, including those with zero gated advantage.
Dual clipping only limits extreme ratio contributions; which student-derived directions receive supervision is still decided by the teacher gate.

\subsection{Compatibility with the MiMo Teacher}
\label{app:mimo-compatibility}
When distilling MiMo-7B-RL-0530 into Qwen3-1.7B, we align the response vocabulary and the prompt text while keeping each model's own prompt tokenization.

\paragraph{Shared response vocabulary.}
The MiMo and Qwen3 output heads have 151,680 and 151,936 rows, respectively.
Token IDs 0--151,664 are identical in the two tokenizers: they cover the 151,643 BPE tokens and the 22 special tokens from \texttt{<|endoftext|>} to \texttt{<|file\_sep|>}, with the same normalization, pre-tokenization, and decoding.
The vocabularies differ only beyond this range.
Qwen3 adds four special tokens that MiMo lacks, \texttt{<tool\_response>}, \texttt{</tool\_response>}, \texttt{<think>}, and \texttt{</think>} (IDs 151,665--151,668), and the remaining rows of both heads (MiMo IDs 151,665--151,679 and Qwen3 IDs 151,669--151,935) are padding with no assigned token.
We therefore restrict responses to the 151,665 shared IDs: a custom SGLang logit processor sets all other logits to $-\infty$ before sampling, and the old and updated student probabilities, policy entropy, and teacher and base scores are all normalized over this shared vocabulary.
Excluding \texttt{<think>} and \texttt{</think>} from responses matches the non-thinking template, in which these tags appear only in the prompt.
Model weights and output heads keep their original dimensions.

\paragraph{Prompt alignment.}
The common ChatML text appends \texttt{/no\_think} to the user message and retains Qwen3's empty thinking block in the assistant prefix.
The student and base receive the original Qwen3 prompt IDs, including the prompt-only thinking tags.
For MiMo, we decode the entire prompt while preserving special tokens and whitespace, then re-encode it with the teacher tokenizer.
We append the sampled response IDs unchanged and align causal scores using the teacher's re-encoded prompt length.
This preserves the prompt text and response targets even when the two tokenizers produce different prompt lengths.
After conversion, we check that the prompt and response fit within the teacher's context length.

\paragraph{Scoring and optimization.}
We run the frozen MiMo teacher with its own Transformers model code and score the student responses by teacher forcing.
OPD uses the sampled teacher--student log-ratio without centering or a base model.
ExOPD uses the same teacher scorer, an extrapolation scale of $\lambda=1.25$, and the frozen initial Qwen3-1.7B student as its reference, with all probabilities normalized over the shared vocabulary.
For \SDOPD{}, the current student's top $K=1{,}024$ candidates include the sampled token, and centering uses probabilities normalized over the entire shared vocabulary, with no renormalization within the candidate set.
The student--base coefficient and teacher gate otherwise follow Section~\ref{sec:method}.
In the MiMo setting, OPD, ExOPD, and \SDOPD{} share the vocabulary mask and prompt protocol, and all three use $c_{\mathrm{dual}}=3.0$.

\subsection{ExOPD Formulation and Hyperparameters}
\label{app:exopd}
ExOPD is the reward-extrapolation variant of generalized on-policy distillation \citep{yang2026learning}.
Our baseline uses the frozen initial post-trained student $S_0$ as its reference, while $p$ denotes the current rollout policy as in Section~\ref{sec:opd-preliminaries}.
In the zero-discount approximation, its token-level advantage is
\begin{equation}
 \begin{aligned}
 A_{\mathrm{ExOPD},t}
 &=\lambda\log\frac{p_T}{p_{S_0}}-\log\frac{p}{p_{S_0}}\\
 &=A^{\mathrm{raw}}_{\mathrm{OPD},t}+(\lambda-1)\log\frac{p_T}{p_{S_0}},
 \end{aligned}
 \label{eq:exopd-feedback}
\end{equation}
where every probability is evaluated at the sampled token $y_t$ and its prefix $h_t$.
The teacher--reference log-ratio supplies an implicit reward, and the student--reference log-ratio is the sampled contribution to KL regularization toward $S_0$.
The scale $\lambda$ controls the reward's weight relative to this regularization.
Setting $\lambda=1$ recovers the OPD advantage in Equation~\ref{eq:opd-feedback}; $\lambda>1$ strengthens the teacher--reference reward, yielding ExOPD.
In the generalized formulation, $0<\lambda<1$ instead corresponds to reward interpolation.

\paragraph{Settings used in this paper.}
We fix $\lambda=1.25$ for every ExOPD configuration, the value used by \citet{yang2026learning}.
The coefficient therefore adds $0.25\log(p_T/p_{S_0})$ to standard OPD while keeping the student--reference term's weight fixed.
The reference $S_0$ remains frozen throughout training and differs from the pretrained base $S_B$ used by \SDOPD{}.
We use $\widehat A_t=\sg(c_{\mathrm{OPD}}A_{\mathrm{ExOPD},t})$ with $c_{\mathrm{OPD}}=1.0$ in the clipped policy loss, without advantage normalization, task rewards, or an additional reference-KL loss.
Shared training hyperparameters are listed in Table~\ref{tab:train-config}; Qwen3 and MiMo runs follow the clipping and scoring settings in Appendices~\ref{app:dual-clipping} and~\ref{app:mimo-compatibility}, respectively.

\section{Supplementary Analysis}
\label{app:analysis}
\setcounter{table}{0}
\setcounter{figure}{0}

\subsection{Full Training Dynamics}
\label{app:sdopd-dynamics}
\begin{figure*}[!t]
 \centering
 \includegraphics[width=0.98\textwidth]{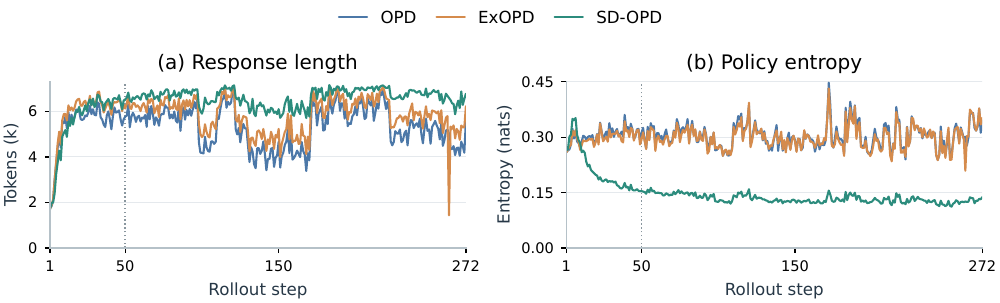}
 \caption{Full training trajectories in the 30B-to-1.7B setting: 272 rollout steps for each method. The dotted line marks step 50, the endpoint of Figure~\ref{fig:training-dynamics}. Each step contains four optimizer updates. Length is the rollout mean; entropy averages the four update means. All recorded points are shown without smoothing.}
 \label{fig:training-details}
\end{figure*}

Figure~\ref{fig:training-details} extends the comparison in Section~\ref{sec:training-dynamics} beyond the first 50 rollout steps.
All three methods use Qwen3-30B-A3B-Instruct-2507 as teacher and Qwen3-1.7B as the initial student.
Each run covers all 272 rollout steps, or 1,088 optimizer updates.

\paragraph{Shared protocol and aggregation.}
All runs use seed 11, DAPO-Math-17K prompts, 64 prompts and four sampled responses per rollout step, and four optimizer updates per step.
The learning rate, sampling settings, and response budget are shared; method-specific clipping follows Appendix~\ref{app:dual-clipping}.
Response length averages all 256 generated responses, including those reaching the 7,168-token cap.
Policy entropy is the arithmetic mean of the four logged update-level entropy means, measured on that method's own response tokens.
The two panels therefore compare generation behavior and predictive concentration under the same rollout schedule.

\paragraph{The early separation persists later in training.}
The first 50 steps capture the rapid initial increase in response length and the emergence of a lower-entropy \SDOPD{} policy.
Later length trajectories continue to fluctuate, while the entropy difference remains clear.
Over the final 20 steps (253--272), mean lengths are 4.73k, 5.25k, and 6.52k tokens for OPD, ExOPD, and \SDOPD{}, respectively; the corresponding entropies are 0.323, 0.322, and 0.128 nats.
For \SDOPD{}, the decrease from 0.158 nats over steps 41--50 to 0.128 over the final 20 steps is gradual relative to its initial decline.
The longer-response, lower-entropy profile of \SDOPD{} thus persists beyond the main-text window, although none of the curves has fully plateaued.

These statistics characterize each policy on the prefixes it generates.
Lower token entropy indicates more concentrated next-token predictions at those prefixes.
Under \SDOPD{}, responses stay long while these predictions become increasingly concentrated.

\subsection{Analysis of Ungated Degeneration}
\label{app:failure}
\paragraph{Observed training dynamics.}
In the 4B-to-1.7B ablation setting, the ungated run gradually collapses into short, low-entropy responses.
Table~\ref{tab:ungated-dynamics} compares the training trajectories of \Ungated{} and \SDOPD{}.
For \Ungated{}, mean response length first falls below 20 tokens at round 105 and reaches six tokens by round 180.
The logged entropy initially rises, reaching 2.1405 at round 60, before falling to 0.0074 at round 120 and 0.0004 at round 272.
The run thus passes through an early phase of growing uncertainty before collapsing onto a few tokens.
\SDOPD{} retains a final training-rollout mean length of 6,176.3 tokens and logged entropy of 0.1243 under the same rollout budget and clipping configuration.

The final evaluation confirms this collapse: none of the \EvaluationResponseCount{} \Ungated{} responses is correct, and all are complete responses of only \UngatedMinLength{}--\UngatedMaxLength{} tokens.
Of these, \UngatedSixTokenCount{} contain exactly six tokens.

\paragraph{Case study: final \Ungated{} outputs.}
The first AIME24 question asks how many Aimeville residents own all four listed items, with a gold answer of $73$.
The first three responses sampled from the final \Ungated{} checkpoint are identical:
\begin{quote}
 \footnotesize\ttfamily
 \detokenize{ShanesPid !_ bureaucr newIndex<|im_end|>}
\end{quote}
Each response contains six tokens, including the termination token, and receives zero reward.
The repeated fragment contains no reasoning about the counting problem and no numerical answer, making the collapse visible in the generated text.

\paragraph{Self-reinforcement without a quality constraint.}
A local policy-gradient analysis relates the student--base signal to divergence from the base.
Fix a prefix and let $\theta_{\mathrm{old}}$ denote the behavior-policy parameters, with $p=p_{S_{\mathrm{old}}}$.
Assume exact full-vocabulary centering and consider the policy gradient at the behavior policy before clipping becomes active.
The base distribution $p_{S_B}$ is fixed, and the sampled coefficient $A_S$ is treated as constant during the update.
Suppressing the fixed prefix in the notation gives
\begin{align}
 G_h &= \left.\mathbb{E}_{a\sim p}\!\left[
 A_S(a)\nabla_\theta\log p_{S_\theta}(a)\right]\right|_{\theta=\theta_{\mathrm{old}}}
 \nonumber\\
 &= \left.\nabla_\theta D_{\mathrm{KL}}\!\left(
 p_{S_\theta}\,\|\,p_{S_B}\right)\right|_{\theta=\theta_{\mathrm{old}}}.
 \label{eq:ungated-kl}
\end{align}
To see this, substitute $A_S(a)=\log[p(a)/p_{S_B}(a)]-D_{\mathrm{KL}}(p\|p_{S_B})$ and use the zero-mean score identity $\mathbb{E}_{a\sim p}[\nabla_\theta\log p_{S_\theta}(a)]=0$ at the behavior policy, as in policy-gradient baseline arguments \citep{sutton1999policy}.
The centering term and the constant term arising when differentiating KL both cancel.
Consequently, at a nonstationary point, a sufficiently small step along this gradient increases divergence from the base.

The signal reinforces tokens whose student--base log-ratio exceeds its current-policy average and suppresses those below it, regardless of answer quality.
Because the numerator is refreshed after each round, newly amplified deviations become part of the next round's supervision.
This feedback can reinforce accidental preferences as well as useful post-training changes.
The identity $D_{\mathrm{KL}}(p\|p_{S_B})=-H(p)-\mathbb{E}_p[\log p_{S_B}]$ also shows why increasing divergence need not preserve entropy: it can favor concentration or movement toward tokens assigned low probability by the base.
Neither direction is intrinsically tied to solving the question.

\paragraph{The role of teacher guidance.}
The teacher gate supplies a direction test that is absent from \Ungated{}: it blocks student coefficients whose signs disagree with centered teacher feedback.
Both methods use the same dual-clipping constant, while task reward, entropy regularization, and auxiliary reference-KL penalties are disabled.
Clipping controls the surrogate incentive for local probability-ratio changes, while the teacher gate selects which student-derived directions receive reinforcement.

Together, the local analysis and the observed trajectories point to a self-reinforcing mechanism: without the gate, deviations amplified in one round become supervision in the next, and unguided preferences grow into degenerate behavior.
The teacher's direction test breaks this loop by admitting only the student changes the teacher also supports.

\begin{table}[t]
 \centering
 \caption{Training dynamics at selected rounds. Length is the mean number of response tokens generated at the start of the round. Entropy is the arithmetic mean of the four logged per-update response-token entropy means. Each method generates its own responses.}
 \label{tab:ungated-dynamics}
 \small
 \setlength{\tabcolsep}{3pt}
 \begin{tabular}{@{}rrrrr@{}}
 \toprule
 & \multicolumn{2}{c}{\Ungated{}} & \multicolumn{2}{c}{\SDOPD{}} \\
 \cmidrule(lr){2-3}\cmidrule(l){4-5}
 Round & Length & Entropy & Length & Entropy \\
 \midrule
1 & 1752.5 & 0.2536 & 1663.9 & 0.2628 \\
20 & 1793.8 & 0.3392 & 5669.3 & 0.1966 \\
40 & 280.6 & 2.0328 & 5880.6 & 0.1636 \\
60 & 1519.9 & 2.1405 & 5777.3 & 0.1469 \\
100 & 30.1 & 0.5400 & 4906.2 & 0.1170 \\
120 & 16.0 & 0.0074 & 6894.2 & 0.1532 \\
180 & 6.0 & 0.0009 & 5665.6 & 0.1195 \\
272 & 6.0 & 0.0004 & 6176.3 & 0.1243 \\
 
 \bottomrule
 \end{tabular}
\end{table}

\section{Token Categorization and Preference Statistics}
\label{app:token-categories}
\setcounter{figure}{0}

\subsection{Observation Protocol}
\label{app:observation-protocol}
We sample 1,024 training questions and analyze all eight student responses per question, without filtering by correctness or length.
Generation uses the non-thinking template, temperature 1.0, top-$p$ 1.0, no top-$k$ truncation, and at most 7,168 response tokens.
The maximum context length is 8,275 tokens, and the per-question generation seed is $11+8\times\text{dataset index}$.
We use vLLM for generation \citep{kwon2023efficient} and Transformers for offline scoring \citep{wolf2020transformers}.
The teacher, student, and base score identical token IDs at the same causal prefixes using untempered logits and full-vocabulary log-softmax.
Log probabilities are computed in FP32 and differenced in FP64; all log probabilities and entropies use natural logarithms.

\begin{samepage}
\subsection{Token Categories}
\label{app:category-definitions}
We assign the five categories in Figure~\ref{fig:amplitude} using fixed English and Chinese phrase lists and textual rules, independently of model scores.

\end{samepage}
\paragraph{Style-related tokens.}
This category covers discourse wording, answer labels, and Markdown formatting used to organize or present a response.
Examples include \emph{let's}, \emph{next}, and \emph{finally}; line-initial \emph{Answer} or \emph{Solution} labels; and Markdown markers such as \verb|#| and \verb|**|.

\paragraph{Mathematical tokens.}
This category covers numbers, operators, formula variables, mathematical vocabulary, and logical expressions used in derivations.
Examples include numerals such as $3$ and $12$, operators such as $+$ and $=$, formula variables such as $x$, and commands such as \verb|\frac| and \verb|\sqrt|.
It also includes terms such as \emph{integer} and \emph{triangle}, and logical expressions such as \emph{if}, \emph{therefore}, and \emph{suppose}.

\paragraph{Markup tokens.}
This category covers formula delimiters, structural punctuation within formulas, and LaTeX commands outside the mathematical and logical command lists.
Examples include \verb|$|, \verb|\[|, and \verb|\]| as delimiters, braces and subscript markers (\verb|_|) within formulas, and formatting commands such as \verb|\text| and \verb|\left|.

\paragraph{Whitespace tokens.}
This category contains tokens composed entirely of whitespace.
Examples include a single space, consecutive spaces, a tab, or one or more line breaks.
Leading spaces attached to a word do not make that token a whitespace token.

\paragraph{Other tokens.}
This category contains unassigned text and punctuation, special tokens, and tokens crossing incompatible category boundaries.
Examples include ordinary prose words such as \emph{the} and \emph{and}, commas or periods in prose, and special tokens such as \verb!<|im_end|>!.

\paragraph{Token boundaries.}
We label character spans and map them to the original tokens through UTF-8 byte spans, without retokenization.
Ignoring whitespace, tokens spanning compatible mathematical or stylistic subtypes keep that category; incompatible mixtures are assigned to Other.
Logical expressions are matched separately and merged into the mathematical group after this boundary handling.

\subsection{Category-Level Aggregation}
\label{app:category-aggregation}
The five displayed categories are disjoint and account for all response tokens in the \ObservationResponseCount{} responses.
For each category, we compute the mean absolute log-ratio within each response, average the eight response means within each question with equal weight, and then average over the 1,024 questions with equal weight.
Every displayed group occurs in every response, so all eight responses contribute to every question's category mean.
For the mathematical group, we combine mathematical-content and logical-expression tokens within each response before computing its mean.
This procedure gives responses equal weight regardless of length.
The student--base and same-sign-filtered panels use exactly the same category positions and averaging order.
For filtering, we multiply each student--base difference by $\ind[r_td_t>0]$; opposite signs and exact zeros yield zero.
Zeroed positions remain in the original category denominator, so the filtered means measure retained amplitude per original token.

\subsection{Expression-Level Preference Statistics}
\label{app:preference-statistics}
We consider alphabetic English strings with at least 50 occurrences across at least 20 questions, merging decoded tokens after lowercasing and stripping leading and trailing whitespace; subword fragments therefore remain separate strings.
The word clouds in Figure~\ref{fig:preferences}(a--b) show 50 representative strings for each sign: ten high-amplitude examples and forty lower-amplitude context strings.
Within each panel, a monotone piecewise-linear size scale expands variation among the context strings while preserving larger sizes for the high-amplitude examples; all strings are horizontal, and color darkens with size.
Figure~\ref{fig:preferences}(c) uses six representative strings chosen by lexical function and observed sign: \emph{suppose}, \emph{try}, \emph{strategy}, and \emph{idea} express hypothesis formation, exploration, and planning, while \emph{straightforward} and \emph{simplicity} describe procedural ease.
They occur in 280, 554, 112, 166, 47, and 110 questions, respectively, and we keep all their occurrences across the eight responses per question.

For each string, we average its occurrences within each response, then the responses containing it within each question, and finally the questions containing it.
We apply this same aggregation to $d_t$, its positive and negative parts, and those parts multiplied by $\ind[r_td_t>0]$.
Rejected positions contribute zero and remain in the original denominators.
The retained amplitude percentage divides the aggregated retained absolute amplitude by the aggregated original absolute amplitude.

\section{Observations across Teachers and Datasets}
\label{app:observation-generalization}
\setcounter{table}{0}

\begin{table}[!t]
 \centering
 \caption{Category-level feedback patterns for Qwen3-1.7B across teachers and datasets. OPD uses teacher--student log-probability differences; student--base uses the student's differences from its pretrained base. Ratios compare mean absolute amplitudes at style-related and mathematical positions, aggregated as in Equation~\ref{eq:amplitude}. Before/After refer to sign gating, which retains positions where the two signals agree; S--B denotes student--base, and Removed gives the share of student--base amplitude removed by gating. $\dagger$For GSM8K, averages use responses and questions containing the category: style-related tokens occur in 8,105 responses across 1,021 questions; mathematical tokens occur in all responses.}
 \label{tab:observation-generalization}
 \footnotesize
 \setlength{\tabcolsep}{3.4pt}
 \begin{tabular}{@{}lrrrrrr@{}}
 \toprule
 & \multicolumn{4}{c}{Style-to-math ratio} & \multicolumn{2}{c}{Removed (\%)} \\
 \cmidrule(lr){2-5}\cmidrule(l){6-7}
 & \multicolumn{2}{c}{Before} & \multicolumn{2}{c}{After} & & \\
 \cmidrule(lr){2-3}\cmidrule(lr){4-5}
 Setting & OPD & S--B & OPD & S--B & Style & Math \\
 \midrule
Original & 3.530 & 4.097 & 3.717 & 3.064 & 69.1 & 58.7 \\
30B-A3B teacher & 3.262 & 4.097 & 3.739 & 3.520 & 63.3 & 57.2 \\
GSM8K data$^{\dagger}$ & 5.646 & 5.246 & 3.811 & 2.318 & 81.0 & 57.1 \\
 
 \bottomrule
 \end{tabular}
\end{table}

For the Qwen3-1.7B student, we examine the category-level amplitude disparity and the effect of teacher selection from Section~\ref{sec:observations} when the teacher or dataset changes.

\paragraph{Controlled settings.}
The original setting uses Qwen3-4B-Instruct-2507 as teacher, Qwen3-1.7B as student, Qwen3-1.7B-Base as base, and DAPO-Math-17K.
The \emph{30B-A3B teacher} setting replaces only the teacher with Qwen3-30B-A3B-Instruct-2507; it reuses the original responses and student/base scores and recomputes teacher scores on exactly the same prefixes and token IDs.
The \emph{GSM8K data} setting retains the original models and samples 1,024 questions without replacement from the GSM8K training split \citep{cobbe2021training}, using the same surrounding prompt instructions.

Each setting includes all eight responses to 1,024 questions, without correctness or length filtering.
Sampling, scoring, and category definitions follow Appendix~\ref{app:token-categories}.
We compare uncentered log-probability differences on the initial student's responses, and teacher-agreement filtering follows Section~\ref{sec:content}, with rejected positions contributing zero to the original averages.

\paragraph{Larger OPD amplitudes at style-related positions.}
Table~\ref{tab:observation-generalization} shows that style-related positions receive larger mean absolute teacher--student log-ratios in both additional settings, with style-to-mathematics amplitude ratios of 3.262 for the 30B-A3B teacher and 5.646 on GSM8K.
Changing the teacher alters the source of feedback on fixed responses, whereas changing the dataset also changes the trajectories being scored.
Both comparisons reproduce the disparity identified in Section~\ref{sec:amplitude}: teacher--student disagreement assigns larger per-token coefficients to differences in expression.

\paragraph{Teacher agreement reduces the relative style amplitude.}
Across the two additional settings, gating removes 63.3--81.0\% of the student--base style amplitude, compared with 57.1--57.2\% of the mathematical amplitude.
The style-to-mathematics ratio consequently falls in each setting, with the largest reduction on GSM8K, from 5.246 to 2.318.
Because filtered positions remain in the original denominators, these reductions measure how much of each category's acquired amplitude survives the teacher's direction test.
Both settings reproduce the observation in Section~\ref{sec:content}: teacher selection changes the relative allocation of student-derived feedback, retaining proportionally less amplitude at style-related positions.

\paragraph{The magnitude source changes the gated balance.}
With the 30B-A3B teacher, gated student--base weighting gives a style-to-mathematics ratio of 3.520, compared with 3.739 for gated OPD.
On GSM8K, the corresponding ratios are 2.318 and 3.811.
Each comparison holds the sign mask fixed, so these differences come from the source of update magnitudes.
Both comparisons reproduce the lower ratio from student-derived magnitudes observed for Qwen3-1.7B in Section~\ref{sec:content}.

\end{document}